\documentclass[conference]{IEEEtran}

\ifCLASSINFOpdf
\else
\fi
\newcommand{\Equref}[1]{Eq.~(\ref{#1})}
\newcommand{\Figref}[1]{Figure~\ref{#1}}
\newcommand{\Secref}[1]{Section~\ref{#1}}
\newcommand{\Tabref}[1]{Table~\ref{#1}}

\usepackage[dvipdfmx]{graphicx}
\graphicspath{%
{./figures/}%
}
\usepackage{amssymb} 
\usepackage{remreset} 

\usepackage{color}

\newcommand{\changeHK}[1]{\textcolor{black}{#1}}
\newcommand{\changeHKK}[1]{\textcolor{black}{#1}}
\newcommand{\changeHKKK}[1]{\textcolor{black}{#1}} 
\newcommand{\changeHKC}[1]{\textcolor{black}{#1}} 
\newcommand{\changeHN}[1]{\textcolor{black}{#1}}
\newcommand{\changeHNK}[1]{\textcolor{black}{#1}} 

\makeatletter
\renewcommand{\thefigure}{%
\arabic{figure}} 
\makeatother

\begin{document}
%
\title{Duration and \changeHN{Interval} Hidden Markov Model for Sequential Data Analysis}

\author{\IEEEauthorblockN{Hiromi Narimatsu}
\IEEEauthorblockA{Graduate School of Information Systems\\
The University of Electro-Communications\\
Chofugaoka 1-5-1, Chofu-shi, Tokyo, 182-8585, Japan\\
Email: narimatsu@appnet.is.uec.ac.jp}
\and
\IEEEauthorblockN{Hiroyuki Kasai}
\IEEEauthorblockA{Graduate School of Information Systems\\
The University of Electro-Communications\\
Chofugaoka 1-5-1, Chofu-shi, Tokyo, 182-8585, Japan\\
Email: kasai@is.uec.ac.jp}}

\maketitle

\begin{abstract}
\changeHK{
Analysis of sequential event data has been recognized as one of the essential tools in data modeling and analysis field. In this paper, after the examination of its technical requirements and issues \changeHK{to \changeHKK{model} complex but practical situation}, we propose a new sequential data model, \changeHN{dubbed} Duration and Interval Hidden Markov Model (DI-HMM), \changeHK{that efficiently represents} ``state duration" and ``state interval" of data events. \changeHK{This has significant implications to play an important role in representing practical time-series sequential data. This eventually provides an efficient and flexible sequential data retrieval.} Numerical experiments on synthetic and real data demonstrate the efficiency and accuracy of the proposed DI-HMM.
} 
\end{abstract}


\IEEEpeerreviewmaketitle

\section{Introduction}
\label{sec:Introduction}

In the context of social science research, \changeHK{the analysis of {\it sequential event data}} has been studied extensively. These studies have explored \changeHK{broad technical areas that range from} biological data analysis to speech recognition, image classification, \changeHK{human} behavior recognition, and time\changeHK{-}series data analysis. Esmaeili {\it et al.} \cite{Esmaeili2010} \changeHKK{categorize three types of} sequential pattern \changeHK{after \changeHKK{theoretical investigation for} large amounts of data}. Lewis {\it et al.} \cite{Lewis1994} propose a sequential algorithm using \changeHK{special} queries \changeHK{to train text classifiers}. Song {\it et al.} \cite{Song2009} propose a sequential clustering algorithm for gene data. More recently, \changeHK{the} studies using sensor data analysis for human behavior recognition and video data understanding have \changeHK{received significant attention} because of the \changeHK{significant progress on} wearable devices \cite{Banaee2013}\cite{Zheng2014}\cite{Cheng2013}. \changeHK{Those} devices enable users to record all of their experiences such as what is viewed, what is heard, and what is noticed. Nevertheless, although \changeHN{collecting} all observed data has become \changeHK{much} easier, it remains difficult to immediately find the data that we want to access  because the amount of time series data is \changeHK{extremely huge}. \changeHK{In case of {\it life log} data application, for example,} it \changeHK{must be much easy} to \changeHK{exactly} retrieve  information of particular places or dates if \changeHK{rich and comprehensive} {\it meta-data} \changeHK{are sufficiently attached to every piece of datum} to \changeHK{be identified}. However, if a query is very ambiguous like {\it retrieving a situation similar to the current situation}, it \changeHK{must be surely} challenging to obtain meaningful \changeHK{results at the end. Thus}, finding \changeHK{such} {\it similar sequential patterns} from vast sequential data using a target pattern extracted from the current situation is of crucial importance. 
\changeHK{This is of interest \changeHKK{in} the present paper}.

\changeHK{Finding similar sequential patterns needs} to discriminate particular sequential patterns from many {\it partial groups} of patterns. There are some useful methods for detecting similar partial patterns from sequential data. One \changeHK{traditional but representative} method is Dynamic Programming (DP) matching \changeHK{algorithm that} is \changeHK{typically} used for \changeHK{speech and} natural language processing. However, because practical sequential data \changeHK{always} contain {\it time misalignments} of events, \changeHK{the sequential pattern} \changeHK{detector must support} a {\it naive} \changeHK{extraction mechanism}, namely \changeHK{the algorithm needs to extract} the sequential \changeHKK{patterns that \changeHN{have}} with \changeHK{not only} the \changeHK{precisely} same duration \changeHK{and} same interval \changeHKK{of events}, but also those with the \changeHK{slightly} different \changeHK{length of} duration \changeHK{and/or the} \changeHK{slightly} different interval. 
%
\changeHK{\changeHK{An alternative} modeling \changeHK{category} is a statistical model of which representatives are} Support Vector Machine (SVM) and Hidden Markov Model (HMM). SVM is introduced in the area of statistical learning and is used for nonlinear classifications such as 
image classification. \changeHK{Although} SVM is \changeHK{powerful to classify data based on the {\it similarity} of each feature}, it is not specialized for sequential data. Meanwhile, HMM is specialized \changeHK{to deal with} sequential data \changeHKK{by} \changeHK{exploiting} {\it transition probability} between states\changeHK{, i.e., events}. Consequently, we \changeHK{dedicate solely to} HMM and its extended methods in this study.

\changeHK{The primary contributions of our work are two-fold:  
(a)} we \changeHKK{advocate} that the support of both ``state duration" and  ``state interval" is \changeHKK{of} great significance to represent practical sequential data \changeHKK{based on an analysis about the} feature and structure of sequential data, \changeHKK{then} extracts requirement\changeHKK{s} for its modeling. \changeHKK{Next}, (b) we propose a new \changeHKK{sequential model} by extending Hidden semi-Markov Model (HSMM) to support both of them \changeHKK{efficiently}. 
Regarding (a), We especially \changeHK{address} the {\it generalization} of the requirements, and \changeHK{emphasize} the importance of handling \changeHK{{\it event order}, {\it continuous ``duration" of an event}, and {\it discontinuous ``interval" time}} between two events. \changeHK{Hereinafter,} we define the event continuous \changeHKK{duration} as ``state duration", and define the discontinuous \changeHKK{interval} \changeHK{with no} observation as ``state interval", respectively \changeHK{because an {\it event} is treated as a {\it state} in HMM.} \changeHK{Then, with respect to (b),} \changeHKK{after assessment of} the extended HMM \changeHK{methods in the literature} \changeHKK{against the requirements}, \changeHK{we show that} all \changeHK{the existing models} cannot treat both \changeHKK{the} state duration and \changeHKK{the} state interval simultaneously. \changeHK{\changeHK{Nevertheless}, we also show that} HSMM, one of the extended HMM methods, \changeHK{can handle the state duration, and is an appropriate baseline to be extended to meet  \changeHK{all the} demands}. 
\changeHK{Subsequently}, in the present paper, we propose an extended \changeHKK{model} of HSMM that accommodate\changeHKK{s} the state duration \changeHKK{as well as} the state interval.

\changeHK{The rest of the paper is organi\changeHN{z}ed as follows.} The next section introduces the related work of sequential data analysis. \Secref{sec:SequentialDataAnalysis} \changeHK{analy\changeHN{z}es} the model requirement\changeHK{, and assesses adequacy} of extended HMM methods \changeHK{against the requirement}. \Secref{sec:DIHMM} explains the proposed method, Duration and Interval HMM (DI-HMM), and \Secref{sec:Experiment} \changeHK{performs} numerical evaluations. Finally, \Secref{sec:Summary} presents a summary of this paper and describes future work.

%
\section{Related Work}
\label{sec:RelatedWork}
This section presents an explanation of related work for sequential data analysis. For sequential pattern matching and detection, DP \changeHK{matching} \cite{Sakoe1978} \changeHK{extracts similar sequential patterns from different two sequential patterns.}  It \changeHK{was} firstly used for acoustic speech recognition, \changeHK{but has been now widely applied to various \changeHN{field}\changeHK{s} such as biological sequences of DNA sequences.} For sequential pattern classification, SVM \cite{Abe2010}
 and Probabilistic Graphical Models \cite{Jensen1990}
 are proposed. SVM is one of the \changeHK{classification algorithms} which produces a model using \changeHK{feature attributes} extracted from training data, and calculates the distance between the training data and test data using these attributes. \changeHK{More recently, however, SVM has been extended to handle} sequential data classification such as speech recognition and handwriting recognition. Shimodaira {\it et al.} propose \changeHK{an} extended SVM which enables frame-synchronous recognition of sequential pattern \cite{Shimodaira2002}. Probabilistic Graphical Model \changeHK{with directed graph or undirected graph} is the typical model to represent sequential patterns. Safari proposes a new model of Deep Learning for sequential pattern recognition \changeHN{\cite{Safari2013}}. \changeHK{Lastly,} HMM \cite{Baum1966}\cite{Eddy1996} is a statistical tool for modeling \changeHN{sequence of observations}. \changeHK{HMM is regarded as one} kind of probabilistic Graphical Model.
%
%
While HMM is used for many applications, for example, speech recognition, handwriting recognition and activity recognition, \changeHKK{most of the} extensions of HMM \changeHKK{are} proposed \changeHK{specialized} for \changeHK{individual} application data.

Considering extraction of not only the \changeHK{exactly} same sequential \changeHKK{pattern} but also the similar sequential \changeHKK{pattern} \changeHK{of time duration and \changeHKK{time} interval of events}, \changeHKK{a} {\it statistical-based} representational capability \changeHKK{is preferably required. To this point of view, although} \changeHK{pattern matching algorithms like DP matching concern {\it sequential oder of events}, \changeHKK{they} do not \changeHK{address} finding such similar sequential pattern\changeHKK{s} due to the lack of  statistical modeling capability}. \changeHK{Furthermore, because} most of statistical \changeHK{modeling algorithms such as} SVM \changeHK{\changeHK{mainly consider} \changeHK{feature similarities}, \changeHKK{they} do not \changeHKK{also} \changeHK{directly} take into account the structure of sequential data. \changeHK{Hence}, they do not handle {\it temporal order and ambiguity} of events to find such similar sequential patterns \changeHKK{efficiently}.}
On the other hand, HMM, a statistical model, is powerful to treat the sequential \changeHKK{patterns} \changeHK{by exploiting}
probability of \changeHK{event} order\changeHKK{s} by transition probability between states. Therefore, we \changeHK{conclude that HMM is the most suitable model to treat the sequential data, and has potential capability to describe temporal ambiguity} to find the similar sequential \changeHKK{patterns}. As a result, we particularly examine HMM \changeHK{hereafter} in \changeHKK{this paper}.

Many extended HMM methods \changeHK{have been} proposed for respective application data. Some of the extended HMM is specialized for sequential data. 
Xue {\it et al.} propose transition-emitting HMMs (TE-HMMs) and state-emitting HMMs (SE-HMMs) for treat\changeHN{ing} discontinuous symbol\changeHKK{s} \cite{Xue2006}. Their studies are for \changeHK{an} off-line handwriting word recognition, and the observation data include discontinuous and continuous \changeHN{parts} between characters when writing cursive letters. They \changeHK{address} such discontinuous and continuous features, and extend HMM to treat both of them. 
\changeHN{Bengio {\it et al.} propose IO-HMM for gesture recognition \changeHK{that maps} \changeHN{ input sequences to output sequences \changeHK{during} learning \changeHK{whereas the} original HMM learn\changeHK{s} only output sequence distribution\changeHKK{s}} \cite{Bengio1995}. 
\changeHK{IO-HMM} supports \changeHN{\changeHK{a} new function \changeHK{of} maximum likelihood and parameters for calculating the maximum likelihood extracted from training data which is the pair of input/output sequences}. 
\changeHK{IO-HMM} is a hybrid \changeHK{model} of generative and discriminative models to treat \changeHKK{output} probability estimation for \changeHK{both of} input sequence\changeHKK{s} and observations.} 
Salzenstein {\it et al.} \changeHN{deal} 
with a statistical model based on Fuzzy Markov random chains for image segmentation in the context of stationary and non-stationary data \cite{Salzenstein2007}. \changeHN{\changeHK{The model handles} multispectral data and \changeHK{estimation of} hyperparameters in non-stationary context}. 
Yu {\it et al.} propose Explicit-Duration Hidden Markov Model \changeHN{\cite{Yu2003}}. 
\changeHN{\changeHK{Addressing} \changeHKK{state} interval between state transition, \changeHK{a} new forward-backward algorithm to estimate  model parameters \changeHK{is proposed}. \changeHN{The model treats the difference of \changeHKK{state} duration\changeHKK{s} in all the states. Beal {\it et al.} propose HMM-selftrans \changeHK{that} is \changeHK{an} extended model of EDM \cite{Beal2002}. Furthermore,} Yu {\it et al.} \cite{Yu2010} and Murphy {\it et al.}  \cite{Murphy2002} propose HSMM which is \changeHKK{a} basic model of EDM. \changeHK{Their new model treats} \changeHKK{the state} duration and the number of observations being produced while \changeHKK{staying} in the state. \changeHN{HSMM is \changeHK{applicable to many applications such as} handwriting recognition, human behavior recognition, and other time series data application estimation \cite{Yu2003MobilityTracking}\cite{Park1996}\cite{Yoma2002}. \changeHK{Hen\changeHN{c}e,} these application data \changeHK{are} especially sequential.} }
Although many extended HMM methods exist, they lack some capabilities to efficiently handle sequential data as the next section explains, \changeHK{more specifically}, the capability to handle both \changeHKK{the state} duration and \changeHKK{the state} interval between events.

%
\section{Sequential Data Analysis}
\label{sec:SequentialDataAnalysis}
This section presents a description of the model requirements for sequential data analysis, and presents comparison of the requirements and the satisfaction of each extended HMM.
%
%
\subsection{Notations}
\label{sec:Notation}

The symbols and the marks are defined {\it a priori} in this section. 
\changeHKK{First, t}\changeHN{aking a certain time as $t$, we \changeHKK{consider} the sequence \changeHKK{of which period is} \changeHNK{$1 \leq t \leq T$}.}
\changeHN{The observation \changeHKK{at} time $t$ is represented as $o_t$\changeHKK{,} and the }\changeHNK{observation sequence starting at time $t=t_1$ to $t=t_2$ is represented as ${\bf o}_{\it {t_{\rm 1}:t_{\rm 2}}}=o_{t_1}, \cdots, o_{t_2}$. 
\changeHKK{The length of ${\bf o}_{\it {t_{\rm 1}:t_{\rm 2}}}$, i.e., $|{\bf o}_{\it {t_{\rm 1}:t_{\rm 2}}}|$, is $t_2-t_1+1$.}
Then, the target observation sequence, which has to be assigned a meaningful label} \changeHKK{to,}
starting at time $t$=1 \changeHKK{and ending} at $t=T$ is represented as ${\bf o}_{\it {{\rm 1}:T}}=o_1,\cdots,o_T$, 
and the set of observable values is $V=\{v_1, \cdots, v_K\}$. 
\changeHKK{Next, an elemental state is denoted as \changeHKKK{$s_{elm}$}, and each state \changeHKKK{$s_{elm}$} has a different length of time units defined as \changeHKKK{$d_{elm}$}. In addition, each has one hidden state that belongs to t}he set of hidden states \changeHKK{denoted as} \changeHN{$S=\{S_1,\cdots,S_M\}$}\changeHKK{, where the size of the set, i.e., $|S|$, is $M$. 
Furthermore, \changeHKKK{$s_{elm}$} is denoted alternatively by using its starting time $t_1$ and its ending time $t_2$, as $s_{t_1:t_2}$. In this case, the time length of $s_{t_1:t_2}$ is equals to $t_2-t_1+1$. On the other hand, a state sequence is generated from multiple elemental states. This state sequence is represented as ${\bf s}_{\it {t_{\rm 1}:t_{\rm 2}}}$ in a bold fon\changeHKKK{t} by specifying its starting time $t_1$ and its ending time $t_2$. \changeHKKK{The $n$-th constitutional state in $s_{t_1:t_2}$ is denoted as $s_n$ of which duration is $d_n$. 
\changeHKKK{For instance, the first state is represented as $s_1$; the duration of $s_1$ is $d_1$.}
In particular, t}he state sequence corresponding to the entire period to be consider\changeHKKK{ed}, namely from time $t=1$ to $t=T$, is denoted as 
${\bf s}_{1:T}=s_1,\cdots,s_N$, where $N$ is the total number of $s_i$. It should be noted that the definition of the} \changeHNK{state sequence is similar to \changeHKK{that of} the observation sequence, but the number of \changeHKK{constitutional} states, \changeHKK{i.e., $|{\bf s}_{1:T}|$,  is not equals to $T$ because each state may have a different time length as explained above.}}

\subsection{Requirement for Model Description}
\label{sec:RequirementForModel}

\begin{figure}[htbp]
\begin{center}
\includegraphics[width=\columnwidth]{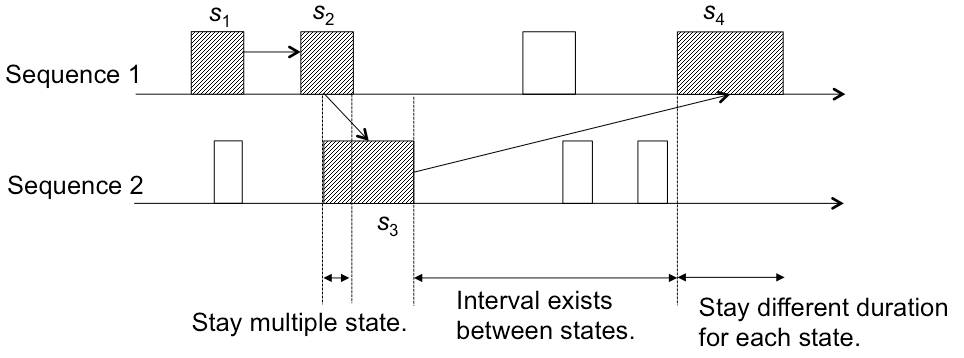}
\caption{Sequential data analysis.}\label{fig:RequirementForTheModel}
\end{center}
\end{figure}

This section presents discussion of the requirements for the \changeHKK{sequential data} model \changeHKK{by} using time\changeHKK{-}series data: representative data of sequential data, as shown in \Figref{fig:RequirementForTheModel}. \changeHK{Suppose that two different sequences \changeHKK{where successive states, i.e., events,} are \changeHKK{observed} from two different sensors. }
\changeHKK{A state is represented as a block of which width represents its continuous state \changeHKK{duration}}.
\changeHKK{Furthermore, those events are not continuously observed, that is, a discontinuous interval time between two observed states may exist.}
\changeHKK{Thus,} the length of this unobserved \changeHKK{period} is represented by \changeHK{distance} between two \changeHK{successive} blocks.
\changeHK{On the other hand}, the \changeHK{gr\changeHNK{a}y-colored} state in \Figref{fig:RequirementForTheModel} \changeHKK{illustrates} the extracted \changeHNK{states sequence, ${\bf s}_{1:T}$, that forms one} group. \changeHKK{The} group of \changeHN{	state sequence, ${\bf s}_{1:T}$,} \changeHK{in this figure} consists of four states: \changeHN{$s_1$, $s_2$, $s_3$, and $s_4$}. 


\changeHK{\changeHKK{Taking into account of the example above, we now} investigate the requirements for the sequential data model}.
\changeHK{First, it goes without saying that the states in \Figref{fig:RequirementForTheModel}} are observed in a prescribed order. \changeHK{This must be described in the model (R1).  Next, in some cases, several} states \changeHK{can} be observed in a partially overlapped manner as \changeHN{$s_2$ and $s_3$}. In other words, multiple states might occur simultaneously at a certain \changeHK{period}. \changeHK{Therefore, such overlapped multiple states must be represented in the model}  (R2).
\changeHK{Furthermore}, the time lengths of respective states mutually differ. \changeHK{This requires the model} to express a `state duration' in a model (R3). Finally, for the case in which \changeHK{two states do not occur in a series without time gap}, a vacant time between one state and another state that is not involved in the group of sequence might exist between two states. \changeHK{Additionally}, \changeHK{the time length of} this vacant time can be variable. Therefore, the `state interval' between two states in the model must be described (R4). Consequently, \changeHK{we conclude} the \changeHK{sequential data} model need\changeHKK{s} to describe the items below.

\begin{itemize}
 \item[R1] State \changeHK{o}rder
 \item[R2] Staying multiple states in a certain time
 \item[R3] State \changeHK{d}uration
 \item[R4] State \changeHK{i}nterval
\end{itemize}

\changeHK{It should be noted that R2 has a different characteristic} from other items because R1, R3, and R4 are the requirements for a  \changeHK{single} sequence \changeHK{whereas} R2 is \changeHK{particular for} multiple sequences. Therefore, we specially examine requirements R1, R3, and R4 in this study. The examination of R2 shall be left for advanced research to be undertaken in future work.


\subsection{Requirement \changeHKK{Assessment} for Extended HMM Methods}  
\label{sec:RequirementVerification}

This section \changeHK{assesses whether} HMM and extended HMM methods meet the requirements \changeHK{analyzed in the previous section.}
\Tabref{tab:Satisfaction} presents a comparison of the conventional HMM methods from the viewpoints of the model requirements. The \changeHK{basic} HMM represents the order of the states, \changeHK{and} all the extended HMM methods inherit this capability. 
\changeHN{IO-HMM treats the state duration but \changeHN{the target of estimation is different}. 
Therefore, it does not satisfy  \changeHKK{the} model requirements.} HSMM is proposed to model \changeHN{the remaining time length to stay} in the same state. In addition, HMM-selftrans and EDM are the extended models of HSMM. These methods satisfy the same requirements: the state order and the state duration\changeHKK{, but do not support the state interval}. As a result of requirement \changeHKK{assessment}, \changeHK{we find that} no model can accommodate both the state duration and the state interval simultaneously. 
\changeHK{Nevertheless, HSMM express\changeHN{es} \changeHKK{the} \changeHN{state} duration, and this capability is not \changeHK{supported} by \changeHK{any} other methods. Therefore, we conclude that} HSMM is the best starting model for extension to our new model. 
The next subsection explains the general HSMM.


\begin{table}[htb]
\small 
\begin{center}
  \caption{Requirement \changeHKK{assessment} of each method.}
  \begin{tabular}[width=\columnwidth]{l|c|c}\hline 
  \shortstack{}&\multicolumn{2}{|c}{Requirements} \\ 
  \hline 
  \shortstack{Method}&\shortstack{\changeHKK{State duration}}&\shortstack{\changeHKK{State interval}} \\ 
  \hline 
  \hline 
  HMM \cite{Eddy1996}&& \\ \hline
  IO-HMM \cite{Bengio1995}&& \\ \hline
  HSMM \cite{Yu2010}\cite{Murphy2002}&\checkmark& \\ \hline
  HMM-selftrans \cite{Xue2006} &\checkmark& \\ \hline
  EDM \cite{Yu2003}&\checkmark& \\ \hline
  DI-HMM \changeHK{(Proposal)}&\checkmark&\checkmark \\ \hline
  \end{tabular}
  \label{tab:Satisfaction}
\end{center}
\end{table}

%

\section{Hidden semi-Markov Model (HSMM)}
\label{sec:HSMM}

\begin{figure}[htbp]
\begin{center}
\includegraphics[width=\columnwidth]{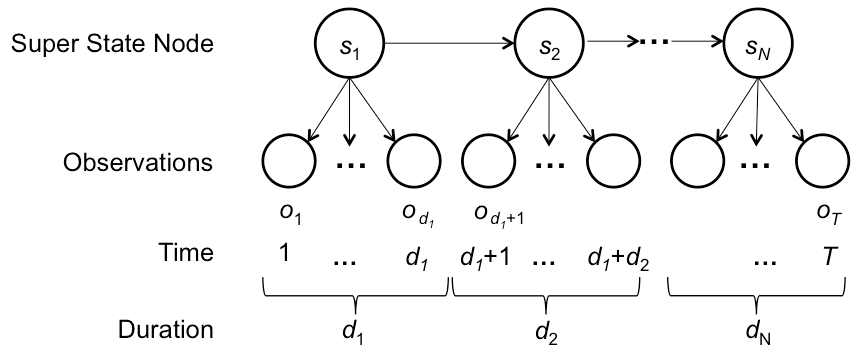}
\caption{State duration handling by HSMM.}\label{fig:ConceptOfHSMM}
\end{center}
\end{figure}

\changeHN{HSMM is an extended model of conducting HMM using a semi-Markov chain with a variable staying duration for each state \cite{Yu2003}. The \changeHN{crucial} difference between HMM and HSMM is the number of observations per state.} HSMM treats the duration of staying at \changeHK{one} state by \changeHK{introducing an} additional parameter \changeHKK{\changeHN{specialized} for describing the state} duration \changeHK{when} calculating the transition and emission probabilities. \Figref{fig:ConceptOfHSMM} \changeHK{illustrates} the concept of HSMM to handle the \changeHKK{state} duration in each state \changeHKK{when} \changeHN{$s_1$, $s_2$, and $s_{\changeHNK{N}}$} are the super state node\changeHK{s}.
\changeHKK{Suppose that the state duration of  \changeHN{$s_1$ is $d_1$, and $\{o_{1}, o_{2}, \cdots, o_{d_1}\}$ is the set of emitted observations from $s_1$ during $d_1$. }} After the state duration time $d_1$ is expired,  \changeHN{$s_1$} is transmitted to the next state  \changeHN{$s_2$}. \changeHK{Thus, HSMM handles the state duration time in each state \changeHN{$s_{\changeHNK{n}}$} by $d_{\changeHNK{n}}$. }

\changeHNK{Here, let $\changeHNK{S_{\it {m'}}}$ and $\changeHNK{S_{\it {m}}}$ be hidden states, and $\changeHNK{D_{m'}}$ and $ \changeHNK{D_{m}}$ be the lengths of time spent in states $\changeHNK{S_{\it {m'}}}$ and $\changeHNK{S_{\it {m}}}$, \changeHK{respectively}. 
\changeHNK{Therefore, $S_{\it {m'}}$ and $S_{\it {m}}$ may happen more than once in a sequence, but we use the same $D_{m'}$ and $D_{m}$ regardless of the number of occurrences in the following equations. }
The number of observations from the state \changeHNK{$S_m$} is determined by the the duration $ \changeHNK{D_{m}}$.} \changeHNK{ \changeHKKK{Thus}, the time length of the observation sequence is calculated as $T=\Sigma^{\changeHKK{N}}_{n=1}d_n$.}
%
%
%
\changeHKKK{Then, t}he transition probability from \changeHN{$\changeHNK{S_{m'}}$} with duration \changeHN{$\changeHNK{D_{m'}}$ to $\changeHNK{S_{m}}$ with duration $ \changeHNK{D_{m}}$ is represented as  $a_{(\changeHNK{S_{m'}},\changeHNK{D_{m'}})(\changeHNK{S_{m}}, \changeHNK{D_{m}})}$.} \changeHKK{This  is defined as}
%
%
%
\changeHNK{
\begin{eqnarray}\label{eq:eq1}
a_{(\changeHNK{S_{m'}},\changeHNK{D_{m'}})(\changeHNK{S_{m}}, \changeHNK{D_{m}})} 
 \triangleq  P[s_{t:t+ \changeHNK{D_{m}}\!-\!1}\!=\! \changeHNK{S_{m}} | s_{t-\changeHNK{D_{m'}}:t\!-\!1}\!=\!\changeHNK{S_{m'}}]. 
\end{eqnarray}
}
%
%
%
The emission probability  $b_{\changeHNK{S_{m}}, \changeHNK{D_{m}}}({\bf o}_{t:t+ \changeHNK{D_{m}}-1})$ is denoted as
%
%
\changeHN{
\begin{eqnarray}\label{eq:eq2}
b_{\changeHNK{S_{m}}, \changeHNK{D_{m}}}({\bf o}_{t:t+ \changeHNK{D_{m}}-1}) \triangleq P[{\bf o}_{t:t+ \changeHNK{D_{m}}-1} | s_{t:t+ \changeHNK{D_{m}}\changeHKC{-}1}=\changeHNK{S_{m}}].
\end{eqnarray}
}
\changeHKK{Finally, t}he set of HSMM parameters, $\lambda$, is defined as
%
%
\begin{eqnarray}\label{eq:eq3}
\lambda  \triangleq  \{a_{(\changeHNK{S_{m'}},\changeHNK{D_{m'}})(\changeHNK{S_{m}},\changeHNK{D_{m}})}, b_{\changeHNK{S_{m}},\changeHNK{D_{m}}}(\changeHKC{{\bf v}_{k_{1}: k_{D_{m}}}}),\pi_{\changeHNK{S_{m'}}, \changeHNK{D_{m}}}\},
\end{eqnarray}
%
where $\pi_{\changeHNK{S_{m'}}, \changeHNK{D_{m}}}$ is the initial distribution of the state $S_{m'}$, 
and ${\bf v}_{k_{1}:k_{D_{m}}}$ represents 
\changeHKKK{the sequence of the observable values of size $D_m$. This is}
$v_{k_{1}}$, $\cdots$, 
$v_{k_{D_{m}}}$ 
$\in V \times \cdots \times V$\changeHKKK{, where $v_{k_n}$ is a $n$-th observable value, and 
$k_n (\leq K)$ is the corresponding index in $V$.}
%
%
%
For the estimation of the likelihood probability, we use an extended Viterbi algorithm \cite{Forney1973} because it is the most popular algorithm for estimating the maximum likelihood. The forward variable in the algorithm is defined as
%
%
%
\begin{eqnarray}\label{eq:eq4}
\delta_t(\changeHNK{S_{m}}, \changeHNK{D_{m}})
&\triangleq & \max_{s_{1:t-d}}P[s_{1:t- \changeHNK{D_{m}}},\changeHNK{s_{t- \changeHNK{D_{m}}+1:t}}=\changeHNK{S_{m}},{\bf o}_{1: \changeHNK{D_{m}}}|\lambda] \nonumber \\
&=&\max_{\changeHNK{S_{m'}} \in S \backslash \{\changeHNK{S_{m}}\}, \changeHNK{D_{m'}}} \{\delta_{t- \changeHNK{D_{m}}}(\changeHNK{S_{m'}},\changeHNK{D_{m'}})\cdot \nonumber \\
&& a_{(\changeHNK{S_{m'}},\changeHNK{D_{m'}})(\changeHNK{S_{m}}, \changeHNK{D_{m}})}\cdot b_{\changeHNK{S_{m}}, \changeHNK{D_{m}}}({\bf o}_{t- \changeHNK{D_{m}}:t-1}\},
\end{eqnarray}
%
%
%
where, \changeHN{$\delta_t(S_{m}, \changeHNK{D_{m}})$} \changeHNK{is} the maximum likelihood that the partial state sequence ends at $t$ in \changeHKK{the} state $S_{m}$ of duration $ \changeHNK{D_{m}}$.
%


%

\section{Duration and Interval HMM (DI-HMM)}
\label{sec:DIHMM}
%
%
\subsection{Motivation}
\label{sec:Motivation}
HSMM introduced in \changeHK{the} previous section handles the state duration. \changeHK{Thus}, HSMM is often called an {\it explicit duration model}. However, \changeHK{because} $S_{m'}$ \changeHKK{moves} to $S_{m}$ \changeHK{at the expiration of duration time $\changeHNK{D_{m'}}$,} HSMM cannot \changeHK{describe} the \changeHKK{state} interval between two states as it is. The easiest way to improve HSMM to handle both the \changeHKK{state} duration and \changeHKK{the state} interval is \changeHK{to introduce} special state, \changeHK{called \changeHN{{\it interval state}},} describing the \changeHN{time} interval between two states. However, it is well known that introducing such \changeHN{an interval state} between two states \changeHK{degrades} the accuracy of discrimination \cite{Natarajan2007}.

\begin{figure}[htbp]
\begin{center}
\includegraphics[width=\columnwidth]{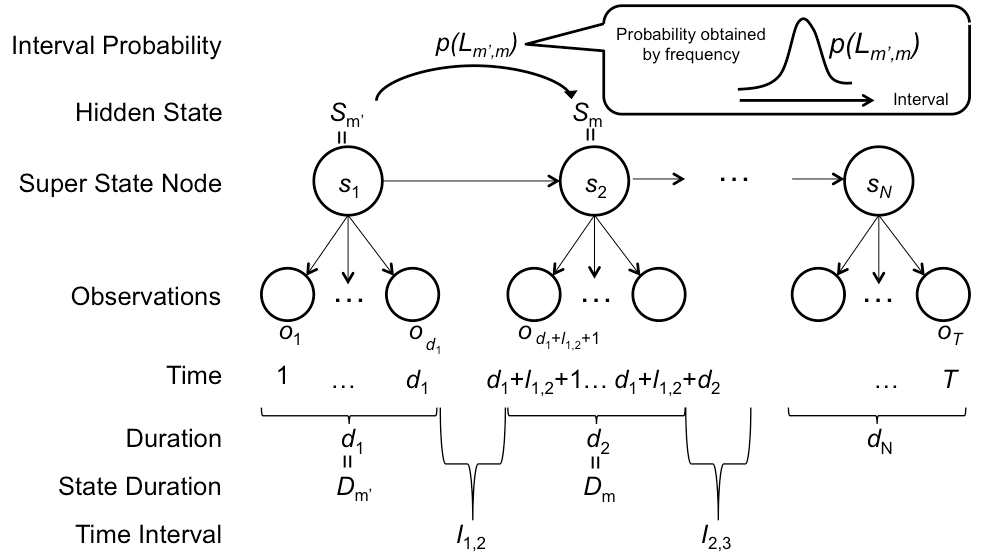}
\caption{Concept of DI-HMM using interval probability.}\label{fig:DIHMM}
\end{center}
\end{figure}

To tackle this problem, we \changeHK{extend the conventional HSMM by newly introducing} {\it \changeHN{state} interval probability} to each transition probability between two states. \changeHK{This present paper calls} this \changeHK{new} extended \changeHK{model} as {\it Duration and Interval Hidden Markov Model} (DI-HMM).
The concept of DI-HMM is \changeHK{illustrated} in \Figref{fig:DIHMM}. \changeHK{Although t}he structure of DI-HMM is similar to HSMM described in \Figref{fig:ConceptOfHSMM}, \changeHN{state interval probability} is newly added \changeHK{to HMM as \changeHKK{\changeHNK{$L_{m',m}$}} as illustrated in the figure}.
\changeHK{While the start time of \changeHN{$S_{m}$} is the next to the end time of \changeHN{$S_{m'}$} in HSMM, \changeHN{$S_{m}$} starts after the \changeHN{\changeHNK{$L_{m',m}$}} length of time passes. 
\changeHNK{The time length of the observation sequence; $T$ \changeHKKK{varies due to its dependency on} the length of the durations and the intervals, \changeHKKK{leading to} $T=\Sigma^{\changeHKK{N}}_{n=1}(d_n+l_{n-1,n})$.} 
\changeHKK{Then, the \changeHN{state} interval} is described by the \changeHN{state interval probability}}. 
\changeHK{Exploiting the \changeHN{state interval probability},} the \changeHK{proposed} model handles the \changeHKK{state} interval in the extended HSMM. The proposed method uses DI-HMM for training a model, and Viterbi algorithm is used for recognition \changeHK{of a test data}.


%
\subsection{Training Sequential Data Model}
\label{sec:ModelingSequentialData}
\begin{figure}[htbp]
\begin{center}
\includegraphics[width=\columnwidth]{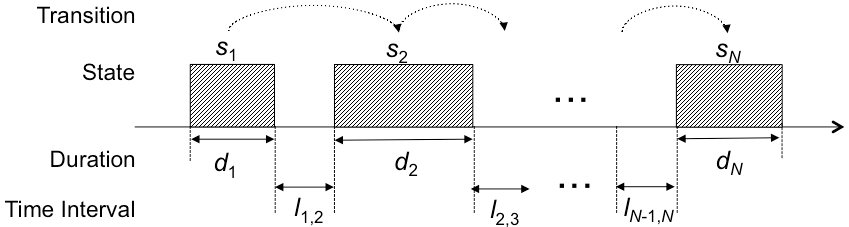}
\caption{Sequential data and representations.}\label{fig:SequentialData}
\end{center}
\end{figure}

The details of DI-HMM is \changeHK{elaborated} using example data \changeHK{as} shown in \Figref{fig:SequentialData}. \changeHN{The slash line patterned blocks} represent the data sequence of training set; 
\changeHNK{$l_{n-1,n}$} is the time \changeHKKK{difference} between the end of $s_{n-1}$ and the \changeHKKK{beginning} of $s_{n}$. \changeHKKK{Furthermore, }
\changeHNK{\changeHNK{$L_{m',m}$}, which is not represented in \Figref{fig:SequentialData}},  \changeHKKK{represents} the time \changeHKKK{difference} between the end of $S_{m'}$ and the \changeHKKK{beginning} of $S_{m}$ \changeHNK{when the next state of $S_{m'}$ is $S_{m}$}. 
%

First, 
the \changeHKK{time} interval probability density distribution is \changeHKKK{expressed} by adopting 
the Gaussian distribution, \changeHN{\changeHNK{$p(L_{m',m})$}}, as
%
%
\begin{eqnarray}\label{eq:eq5}
\changeHNK{p(L_{m',m})}~=~\frac{1}{\sqrt{2\pi\sigma^2}}e^{- { \frac{(x- \mu)^2}{2\sigma^2} }},
\end{eqnarray}
where $\sigma$ donates the variance of \changeHKK{state} interval \changeHN{\changeHNK{$L_{m',m}$}}, and \changeHN{$\mu$} is the \changeHKK{mean} of \changeHN{\changeHNK{$L_{m',m}$}}. 
Then, the set of parameters used in DI-HMM is defined as
%
%
\begin{eqnarray}\label{eq:eq6}
\lambda & \triangleq & \{a_{(S_{m'},\changeHNK{D_{m'}})(S_m, \changeHNK{D_{m}})},b_{(S_{m}, \changeHNK{D_{m}})}
({\bf o}_{1: \changeHNK{D_{m}}}),\nonumber \\
&& \hspace{3.5cm} \pi_{S_{m'}, \changeHNK{D_{m}}}, \changeHNK{p(L_{m',m})}\}.
\end{eqnarray}
%
%

The transition and emission probabilities are defined as being equal to HSMM. The difference between HSMM and DI-HMM is to consider the parameter of \changeHN{\changeHNK{$p(L_{m',m})$}}.
The reason why the Gaussian distribution is adopted as the interval distribution is that it simply expresses the density distribution, and the parameters are not required to change for each probability. But, other distributions and functions for our proposed algorithm could be adopted. 

The range of $x$ might influence either memory consumption \changeHK{and/or} computational complexity to generate the model. There might be no $x$ value suitable for the observation values \changeHKKK{due to the range limitation of $x$}
if \changeHN{\changeHNK{$p(L_{m',m})$}} is generated in a training period. However, if \changeHK{the} parameter \changeHN{\changeHNK{$p(L_{m',m})$}} is generated every time an observation is \changeHK{fed} to the \changeHK{algorithm}, the calculation cost can be much higher. Our motivation to introduce the interval distribution to HSMM is\changeHK{, as explained earlier,} to find the similar part of sequential data including the interval and also to discriminate between \changeHK{the target part} and the similar part. Therefore, even if the probability of \changeHNK{$L_{m',m}$} is presumed to zero around the skirts of the distribution, no particular problem arises. Consequently, we introduce the boundary of the probability value $\theta_{pt}$ to determine the edge of the skirt of \changeHN{\changeHNK{$p(L_{m',m})$}}. On generating the \changeHN{\changeHNK{$p(L_{m',m})$}}, the calculation is terminated when the probability value \changeHK{goes} less than $\theta_{pt}$.


\subsection{Probability Estimation for Recognition}
\label{sec:ProbabilityEstimation}

The Viterbi algorithm is used to estimate the probability of a model as \cite{MurphyThesis2002}. \changeHN{Then, the pair of the model and the probability described are stored as the candidate, and finally} the maximum likelihood estimate is calculated for each state for each model.

First, we calculate the probability of \changeHNK{$L_{m',m}$}, \changeHK{i.e., }\changeHNK{$p(L_{m',m})$}, assigning \changeHNK{$L_{m',m}$} to the new parameter distribution \changeHN{\changeHNK{$p(L_{m',m})$}} beforehand. If \changeHNK{$L_{m',m}$} is out of \changeHK{the} range for the \changeHN{\changeHNK{$p(L_{m',m})$}}, the \changeHN{time} interval probability is determined as
%
%
\begin{eqnarray}\label{eq:eq7}
\changeHNK{p(L_{m',m})} & = & \min_{\changeHKK{1 \leq m' \leq M, 1 \leq m \leq M}} \changeHNK{p(L_{m',m})} \times c. 
\end{eqnarray}
Therein, $c$ is $0 \leq c \leq 1$. 
Then, the forward variable for estimating the maximum likelihood is calculated as
%
%
\begin{eqnarray}\label{eq:eq8}
\delta_t(S_m, \changeHNK{D_{m}}) & \!\! \!\!\!\!  \triangleq \!\! \!\! \!\! & \!\!\!\! \max_{s_{1:t- \changeHNK{D_{m}}}} \!\!P[s_{1:t- \changeHNK{D_{m}}}, 
s_{t-\changeHNK{D_{m'}}+1:t}\!\!=\!\!S_m,{\bf o}_{1: \changeHNK{D_{m}}}| \lambda] \nonumber \\
& \!\!\!\!\!\!    =\!\!\!\! \!\!   & \max_{S_{m'} \in S \backslash \{S_{m}\}, \changeHNK{D_{m'}}} \left\{ \delta_{t- \changeHNK{D_{m}}}(S_{m'},\changeHNK{D_{m'}}) \right. \nonumber \\
&& \left.\cdot 
a_{(S_{m'},\changeHNK{D_{m'}})(S_{m}, \changeHNK{D_{m}})} \cdot b_{S_{m'},\changeHNK{D_{m'}}}({\bf o}_{t-\changeHNK{D_{m'}}+1:t})\right. \nonumber \\
&& \left. \cdot \changeHNK{p(L_{m',m})}\right\}.
\end{eqnarray}
%
%
%

The parameter of \changeHNK{$p(L_{m',m})$} is the same as \changeHKKK{the one}  introduced into \Equref{eq:eq4} in \Secref{sec:HSMM}. The \changeHKKK{state} interval probability is calculated at the same time as calculating the parameter of the likelihood using the transition probability \changeHN{recursively}.

The difference between HSMM and DI-HMM is the capability of \changeHK{handling} the \changeHN{time interval between states} \changeHK{as explained \changeHK{earlier}}. The interval probability \changeHK{in DI-HMM} is integrated for introducing the \changeHN{time interval between states} to calculate the likelihood. This calculation might cause additional calculation cost. \changeHK{Hence}, it is necessary to evaluate the model \changeHKKK{for} calculation cost. In addition, \changeHK{t}he observation distributions \changeHN{$b_{S_{m}, \changeHNK{D_{m}}}({\bf o}_{1: \changeHNK{D_{m}}})$} can be parametric or non-parametric. In this proposal, \changeHN{the relation} of the \changeHN{state} duration and the \changeHN{state} interval is not represented in a model. For that reason, \changeHN{$b_{S_{m}, \changeHNK{D_{m}}}({\bf o}_{1: \changeHNK{D_{m}}})$} is handled as non-parametric, discrete, and independent of the state durations. Then, \changeHN{\changeHNK{$p(L_{m',m})$}} is also discrete and independent of the state duration and the transition probability.

%
\section{\changeHK{Numerical} Experiment\changeHK{s}}
\label{sec:Experiment}

\begin{figure}[htbp]
\begin{center}
\includegraphics[width=\columnwidth]{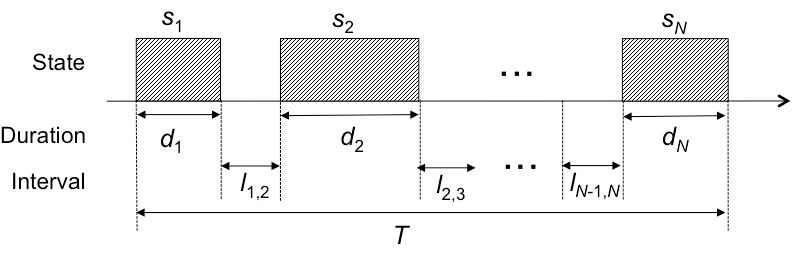}
\caption{\changeHKKK{Sequential data generation. $N$ is given as the number of states.}}\label{fig:DataGenerationPolicy}
\end{center}
\end{figure}

This section presents comparison\changeHKK{s} of the following items \changeHKK{between} DI-HMM and HSMM: section A describes the discrimination performance, section B describes the recognition performance, and section C describes the calculation time of training and recognition of DI-HMM.
\Figref{fig:DataGenerationPolicy} portrays an example of how to generate synthetic data for the evaluation. First, the number of states \changeHNK{$N$}, the minimum value of the duration $d_{min}$, the maximum value of duration $d_{max}$, the minimum value of the \changeHN{state} interval \changeHN{$l_{min}$} and the max value of the \changeHN{state} interval \changeHN{$l_{max}$} are given as initial parameters. The length of sequence \changeHN{$T$} is also given \changeHKKK{in} the evaluation. 
In the example of \Figref{fig:DataGenerationPolicy}, the number of states \changeHKK{and durations are $N$,  and the number of intervals is $\changeHKK{N}-1$}. The lengths of each duration and each \changeHN{state} interval are incremented one by one. Then\changeHKKK{, all those} lengths are combined with round robin.
%
%
%
%
%
%
%
\subsection{Discrimination Performance}
\label{sec:DiscriminationPerformance}

\begin{figure}[htbp]
\begin{center}
\includegraphics[width=0.9\columnwidth]{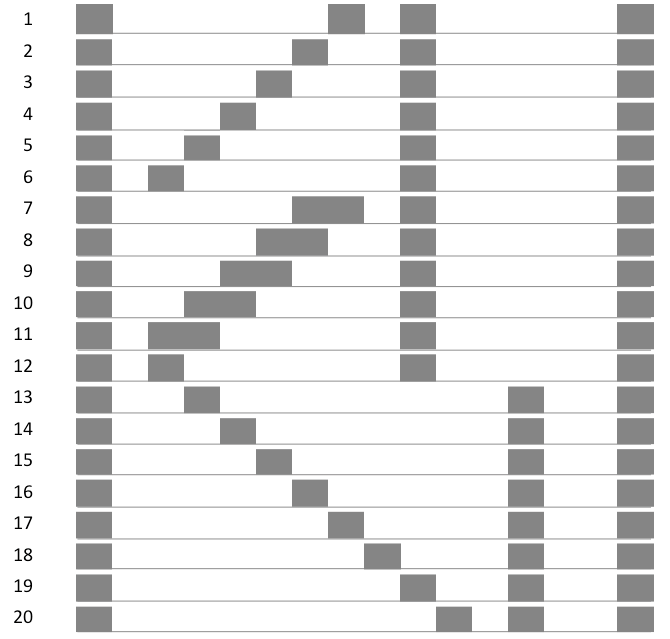}
\caption{Example data sequences from Data 1 through Data 20. }\label{fig:ExampleSequence}
\end{center}
\end{figure}

First, we generate 200 different sequences \changeHKKK{by} fixing \changeHNK{$T=14$}, $\changeHKK{N}=\{3,4\}$, $d_{min}=1$, $d_{max}=10$, \changeHN{$l_{min}=1$, and $l_{max}=4$}. \Figref{fig:ExampleSequence} shows the generated example data. \changeHKK{Each row} represents one sequence of data. 
\changeHNK{The gray blocks} 
are  observed state\changeHKK{s, and} the length of the 
states represents the duration. 
%
%
%
%
%
To evaluate the discrimination performance, we compare the likelihood\changeHKK{s calculated using \Equref{eq:eq8} for} the test data \changeHKK{against} each training datum. \changeHKK{Discrimination means that the 
likelihood for each training data is different from one another}. 
%
%
%
\begin{figure}[htbp]
\begin{center}
\includegraphics[width=\columnwidth]{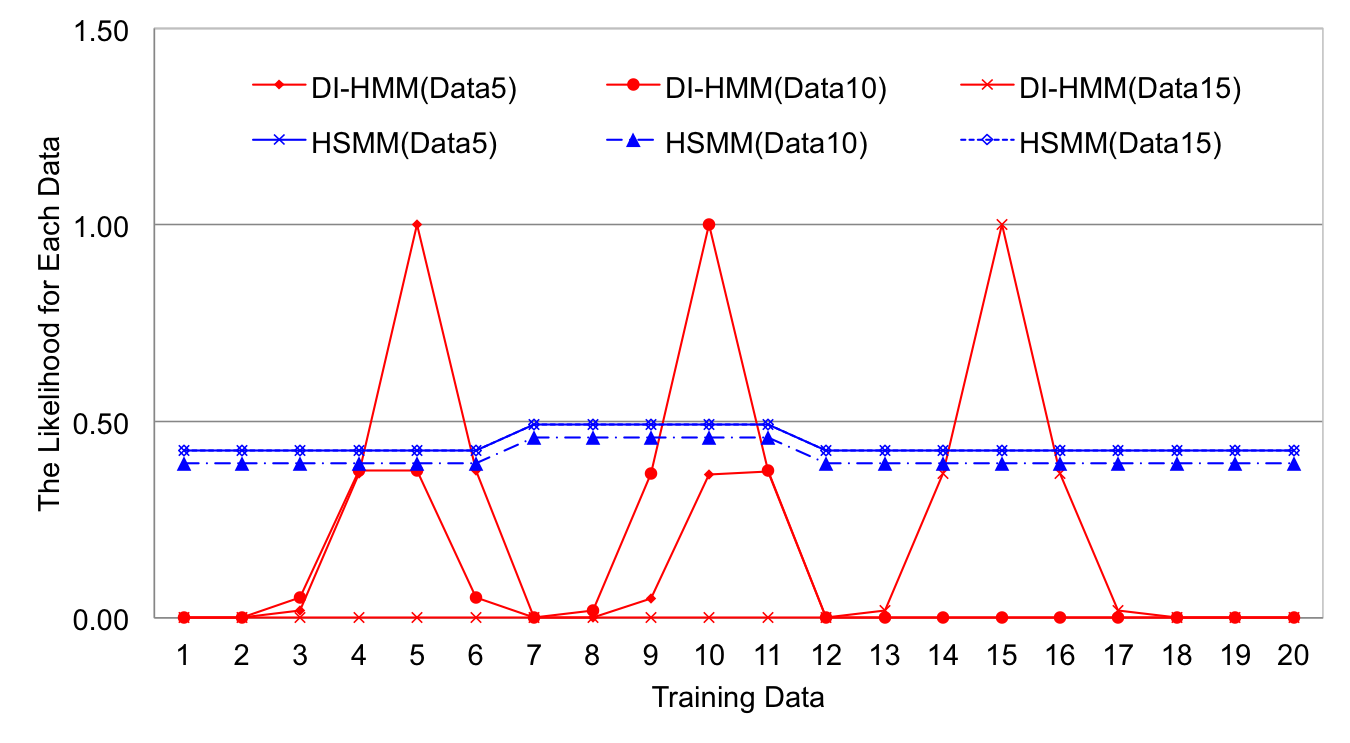}
\caption{Likelihood for respective training data.}\label{fig:LikelihoodOfTrainingData}
\end{center}
\end{figure}

\Figref{fig:LikelihoodOfTrainingData} exhibits the results of Data 5, 10, and 15, extracted by the sequential data presented in \Figref{fig:ExampleSequence}. The $x$-axis shows each training datum\changeHKK{, and} the $y$-axis shows the likelihood for each test \changeHKK{datum}. 
\changeHKK{From this figure,} the likelihood\changeHKK{s} of HSMM \changeHKK{indicate} around 0.5 for each \changeHKK{training} datum; Data 5, 10, \changeHKK{and} 15. 
\changeHKK{This means that all the test data have similar data model, and they cannot be discriminated by HSMM.
On the other hand, in DI-HMM case,} each result has a peak value at the corresponding training data. 
\changeHKK{This means that DI-HMM can discriminate those data, and} DI-HMM can discriminate \changeHN{the differences of both the state duration and the state interval.}

\begin{figure}[htbp]
\begin{center}
\includegraphics[width=\columnwidth]{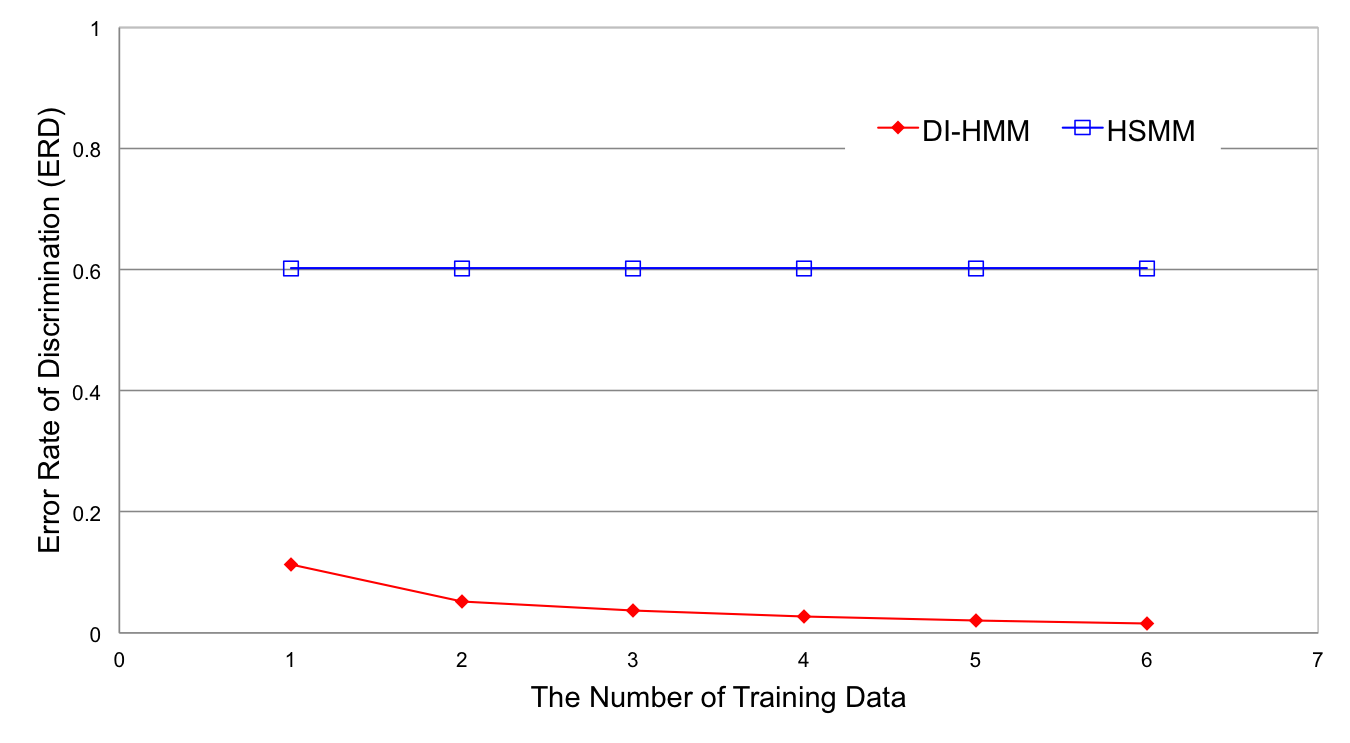}
\caption{\changeHKKK{Number of training data vs. ERD.}}\label{fig:ErrorRateOfDiscrimination}
\end{center}
\end{figure}

\Figref{fig:ErrorRateOfDiscrimination} shows the \changeHKK{discrimination} performance \changeHKK{against} each different dataset \changeHKK{when} changing the number of training data from 1 to 6. The $x$-axis shows the number of training data; the $y$-axis shows the Error Rate of Discrimination ($ERD$). From the result, 
 DI-HMM discriminates between the entire sequences by checking the difference of the \changeHN{state} interval length for the most part. Even if the \changeHK{amount} of training data is only one, the performance is sufficiently high because the $ERD$ is close to 0.1 while the $ERD$ of HSMM is around 0.6 constantly. Therefore, DI-HMM has powerful discriminative capability to recognize differences among sequences.

%
%
%
%
%
\subsection{Recognition Performance}
\label{sec:RecognitionPerformance}
The results of \changeHK{the} discrimination performance show that the likelihood of \changeHKK{DI-HMM} can \changeHK{give} the maximum value \changeHKK{at} the data \changeHKK{that ha\changeHN{s}} the same duration and the same \changeHN{state} interval in all training \changeHKK{data}. However, in a practical field, robustness to a slight time delay or time extension of the same labeled data is required. For instance, \changeHK{in case of} music performance, the \changeHK{consecutive time} lengths of one sound \changeHKK{must be} slightly different  when the same rhythm \changeHKK{is played by} two \changeHKK{different} instruments or played by two players. Therefore, we evaluate the recognition performance \changeHK{when individual sound has such a time \changeHN{delay} 
and/or a different consecutive time.}


\begin{figure}[htbp]
\begin{center}
\includegraphics[width=0.9\columnwidth]{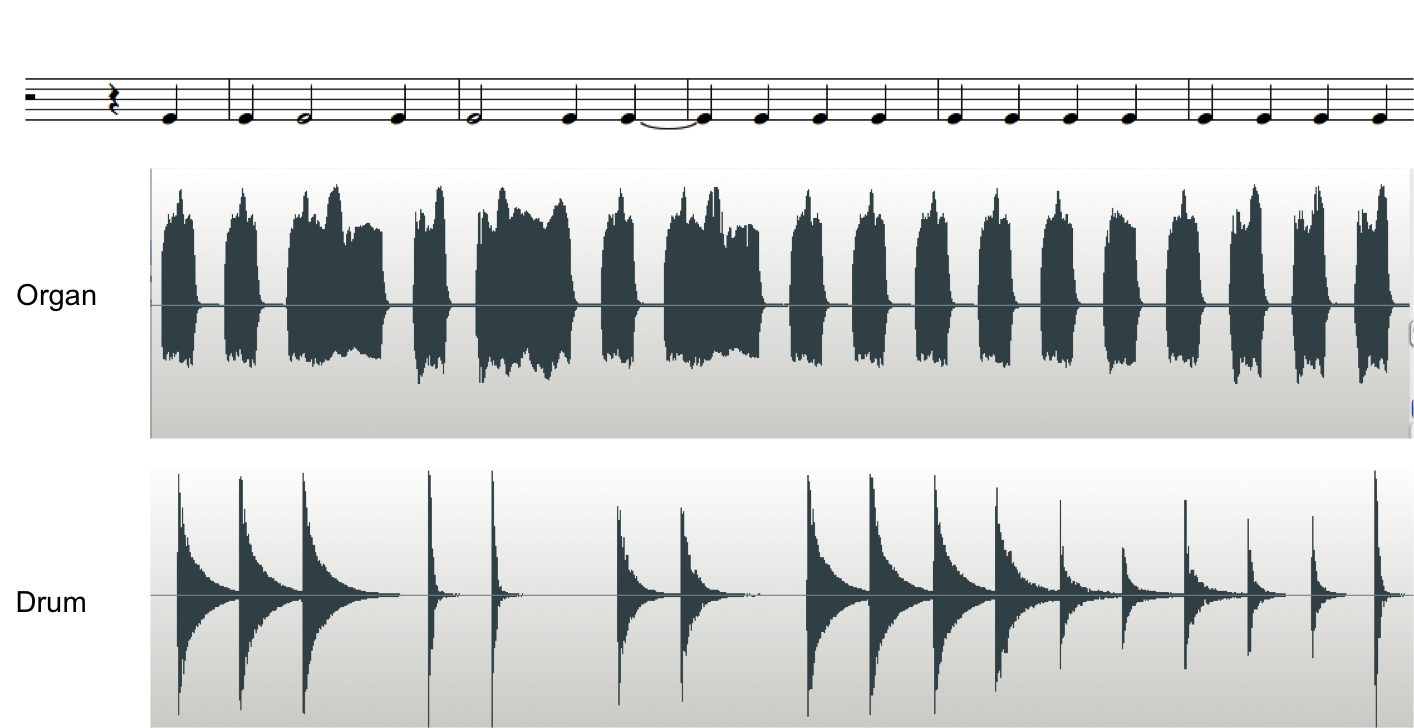}
\caption{Part of an example rhythm score and the waveform.}\label{fig:RhythmScore}
\end{center}
\end{figure}

To evaluate \changeHKK{the impact caused by} the time difference \changeHKK{in} the data assigned \changeHKK{to} the same label, we use music sound data played \changeHKK{by} different instruments. \Figref{fig:RhythmScore} presents an example of the rhythm score\changeHKK{s}. The two waveforms are the monophonic waveform of sound played \changeHKK{by} an organ and a drum. \changeHKK{The waveform of the organ has longer length of notes than the dr\changeHN{u}m has. In this experiment, t}he music data for the evaluation are generated using the following steps:

\begin{itemize}
 \item[a] Divide \changeHK{an input} waveform into bars\changeHKK{, which are small pieces of music containing a fixed number of beats}.
 \item[b] The length of \changeHK{the period when sounds exist} \changeHKK{corresponds to}
 the \changeHKK{state} duration time\changeHKK{, whereas} the length of \changeHK{the period when no sound exists} \changeHKK{corresponds to the state interval time.}
 \item[c] The observation sequence consists of the sound ``on"  symbol and \changeHKK{``off"} symbol.
\end{itemize}

For this experiment, we use the bass part of ``We Wish You A Merry Christmas" \changeHK{that consists} of 57 bars. First, we prepare six \changeHN{kinds of music sound} \changeHKK{which are played by different instruments with different lengths of notes.}
Table II shows \changeHN{each of differences in the experimental data set. }
\changeHKK{In the experiment,} Data 1, 2, and 3 are used as training data; and Data 4, Data 5, and Data 6 are used as test data.


\begin{table}[htb]
\normalsize
\begin{center}
 \caption{\changeHKK{Generated Music Data.}}
  \begin{tabular}[width=\columnwidth]{c|c|c}\hline 
    Index & Instrument & Minimum Length of Note\\ \hline     \hline
    Data 1 & Grand Piano & 3/4 length of crotchet\\ \hline
    Data 2 & Grand Piano & 1/2 length of crotchet\\ \hline
    Data 3 & Puncy Grand Piano & 1/4 length of crotchet\\ \hline
    Data 4 & Electric Piano & 1/4 length of crotchet\\ \hline
    Data 5 & Drum & 1/4 length of crotchet\\ \hline
    Data 6 & Organ & 1/2 length of crotchet\\ \hline
  \end{tabular}
\end{center}
\end{table}

For the training \changeHK{phase}, the first bar\changeHKK{s} of Data 1, 2, and 3 are trained for \changeHK{the} model labeled \changeHKK{as} \changeHK{``1"}. For the recognition \changeHK{phase}, the probability of each model \changeHKK{is calculated}. The recognition result (the estimated label) is \changeHKK{obtained from} the label of the model \changeHKK{with} \changeHK{the} maximum probability. We evaluate the recognition accuracy based on $f$-measure \changeHK{that is calculated by ${2 \cdot {\rm recall} \cdot {\rm precision} /(\rm recall} + {\rm precision})$, where ${\rm precision}=TP/PP$, and ${\rm recall}=TP/AP$. 
Here, when} the Predicted Positive ($PP$) is the number of models whose likelihood calculated using \Equref{eq:eq4} or \Equref{eq:eq8} is the maximum in all models, True Positive ($TP$) is the number of collected models in $PP$; and Actually Positive ($AP$) is the number of labeled models.
We \changeHKK{prepare and} evaluate two patterns of data \changeHKK{by} changing the length of \changeHKK{bars}. \changeHKK{The observation sequence in the first pattern} consists of a sound pattern of \changeHKK{one} bar. \changeHKK{Whereas} each observation sequence consists of a sound pattern combining two continuous bars in the second pattern. \changeHKK{In} the first case, \changeHKK{we obtain 57 labeled models, and extract 40 bars of which their rhythms are different. Thus, these 40 bars finally} are used for the experiment. \changeHKK{At} the training  \changeHK{phase}, 40 models are generated by all 120 (= 40 bars $\times$ 3 data) training data. Then, \changeHKK{at} the recognition \changeHK{phase}, another 120 data are tested. Similarly, \changeHKK{as f}or the second case, the number of the labeled models is 26, \changeHK{and} the number of training data is 78 (= 26 bars $\times$ 3 data). The number of test data \changeHKK{is} 78.

\begin{figure}[htbp]
\begin{center}
\includegraphics[width=\columnwidth]{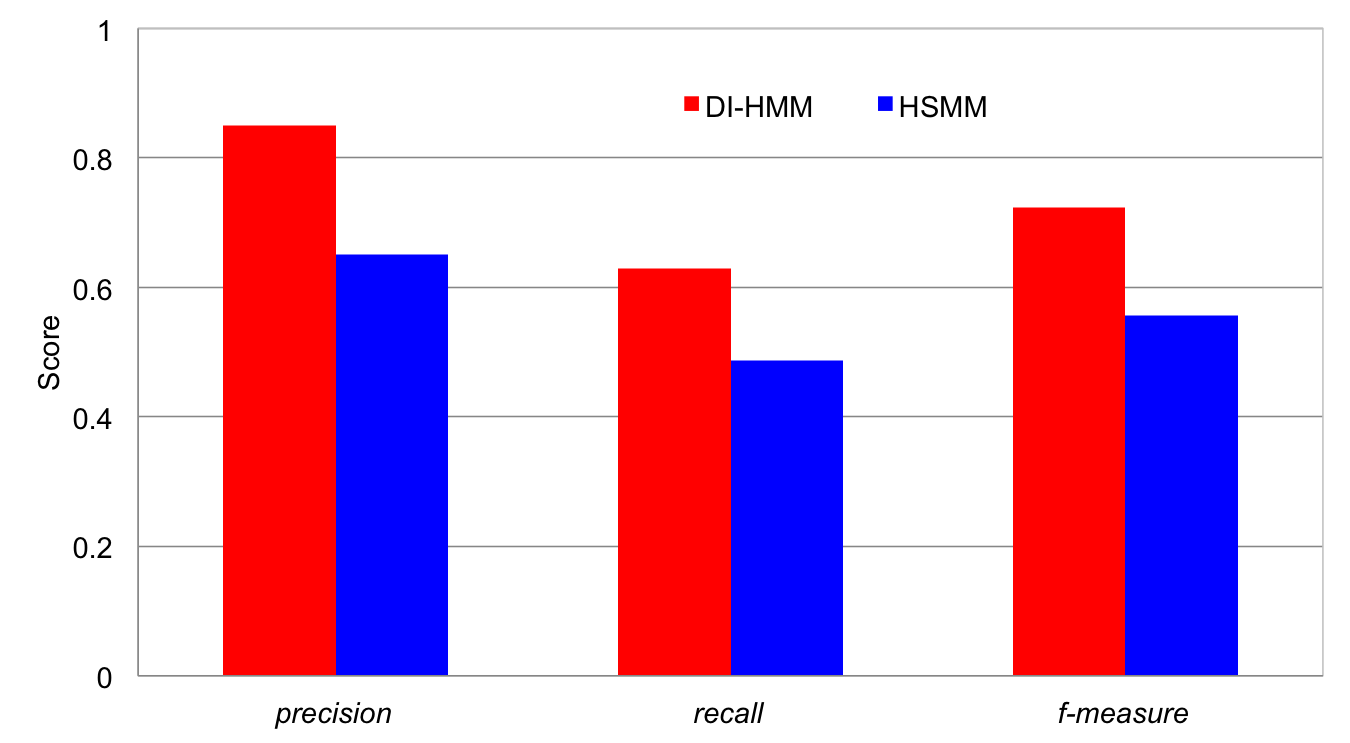}
\caption{Recognition accuracy when the sequence consists of one bar.}\label{fig:RecognitionResultOneBar}
\end{center}
\end{figure}

\begin{figure}[htbp]
\begin{center}
\includegraphics[width=\columnwidth]{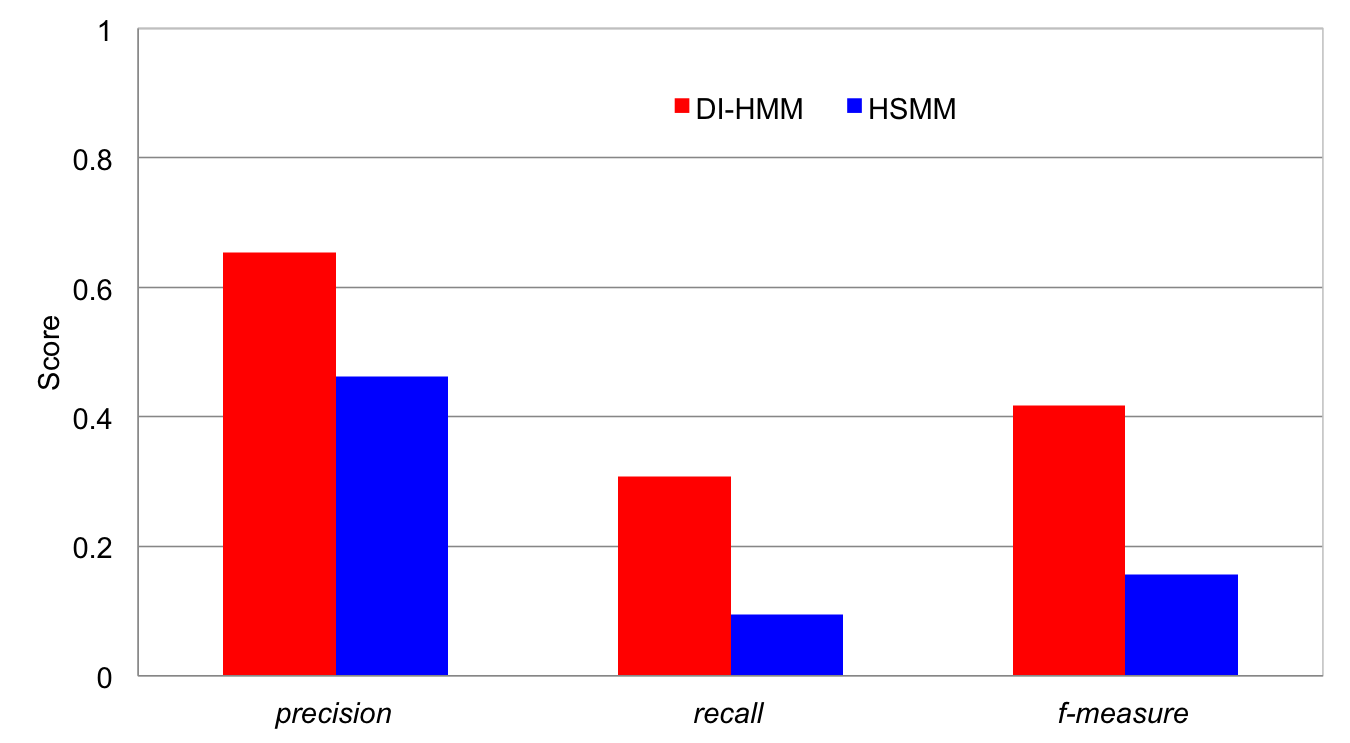}
\caption{Recognition accuracy when the sequence consists of two bars.}\label{fig:RecognitionResultTwoBars}
\end{center}
\end{figure}

\Figref{fig:RecognitionResultOneBar} and \Figref{fig:RecognitionResultTwoBars} show the recognition results in \changeHKK{the} first and second patterns\changeHKK{, where} \changeHN{red bar} graphs 
\changeHKK{show} the results of DI-HMM, and \changeHN{blue bar graphs}
\changeHKK{show} \changeHKK{those} of HSMM. 
\changeHKK{From these figures,} all scores \changeHKK{of} precision, recall, and {\it f}-measure of DI-HMM \changeHKK{indicate} higher \changeHKK{values} than those of HSMM. Therefore, our proposed model is effective for recognition of the rhythm pattern of music \changeHKK{by} taking \changeHKK{into} account various instruments.
\changeHKK{Furtheremore, comparing the results} between \Figref{fig:RecognitionResultOneBar} and \Figref{fig:RecognitionResultTwoBars}, the recognition accuracy for the data of which sequence consists of two bars is worse than that for the data of which each sequence consists of \changeHKK{one} bar. 
\changeHKK{This result can be explained as follows;}
the various \changeHK{lengths of} durations and intervals are trained for the same duration and interval \changeHN{probability} between the same two states \changeHK{when} the length of the sequence is longer and \changeHK{when} \changeHN{the same symbol appears many times in a sequence}. 
However, \changeHK{the} symbols of the observation sequence differ \changeHK{in} practical data like music \changeHK{data} or some life event data. \changeHK{Hence}, the quantities of symbols \changeHK{indicate various different values}. \changeHK{Consequently,} when various symbols are included in a sequence, the recognition accuracy will \changeHK{get increased} even if the sequence is longer.


\subsection{Calculation time of Training and Recognition}
\label{sec:calculation}

\begin{figure}[htbp]
\begin{center}
\includegraphics[width=\columnwidth]{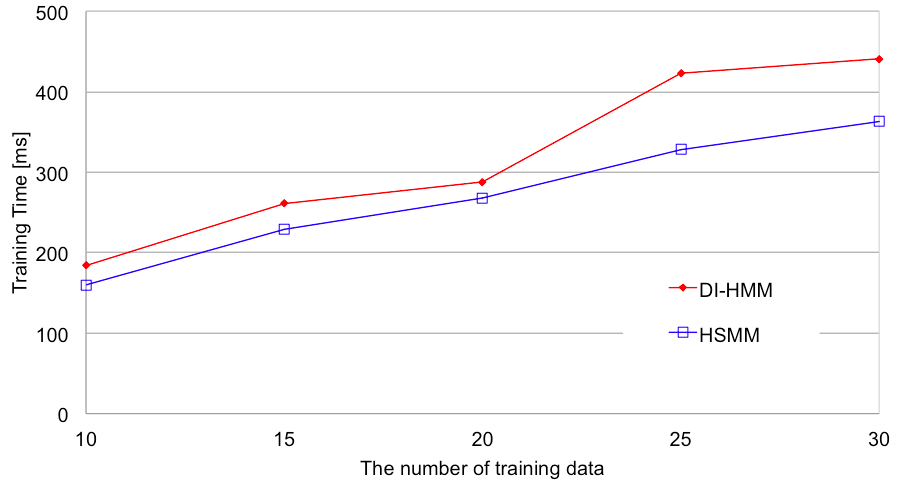}
\caption{Training time with the training dataset size.}\label{fig:TrainingTime}
\end{center}
\end{figure}

\begin{figure}[htbp]
\begin{center}
\includegraphics[width=\columnwidth]{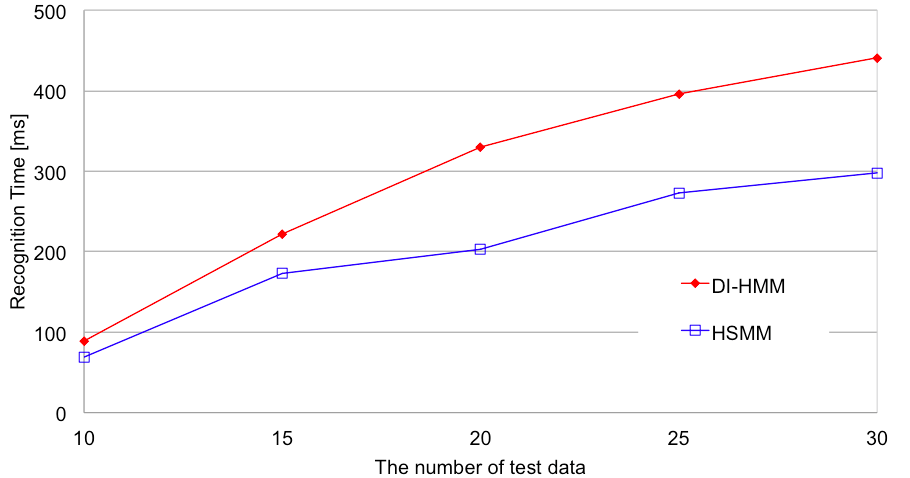}
\caption{Recognition time with \changeHK{test} dataset size.}\label{fig:RecognitionTime}
\end{center}
\end{figure}

For calculation time evaluation, we generate 35 sequences, fixing $d_{min} = d_{max} =2$, \changeHN{$l_{min} =1$, $l_{max} =10$, and $T$} is not fixed {\it a priori}. Using the generated data, we compare training time and recognition time while changing the number of training data.
The results of training time \changeHK{and recognition time are shown in \Figref{fig:TrainingTime} and \Figref{fig:RecognitionTime}, respectively}. The $x$-axis shows the number of training data\changeHKK{, and t}he $y$-axis shows the calculation time for training/recognition. \changeHN{The red line is}
the result of DI-HMM, and \changeHN{the blue line}
is the result of HSMM. 
%
%
The slope\changeHKK{s} of the recognition time and the training time in DI-HMM \changeHKK{are} steeper than \changeHKK{those} of HSMM. Therefore, \changeHK{the} introduction of the interval \changeHN{probability} \changeHK{onto} HSMM is expected  \changeHN{to \changeHK{pose} additional calculation \changeHK{cost}. \changeHKK{However,} it does not \changeHK{severely affect} the total amount of calculation}. \changeHK{Meanwhile, as shown in the previous sections,} the recognition performance\changeHKK{s} and \changeHKK{the} discrimination performance\changeHKK{s} of DI-HMM \changeHK{give superior results} than HSMM \changeHK{does}. Therefore, \changeHK{we conclude that our proposed DI-HMM is very effective for the sequential data analysis that is originally motivated in this paper.}


%

\section{Summary and Future Work}
\label{sec:Summary}

As described herein, \changeHK{we investigated the requirements for sequential data analysis by focusing on} the structure and feature of sequential data. Then, we have proposed Duration and Interval HMM (DI-HMM) by introducing the interval probability onto HSMM \changeHK{in order} to handle both the state duration and \changeHKK{the} state interval. 
We evaluated the discrimination performance, recognition performance, and measured the calculation time for training and recognition by the computational simulation. 
For the evaluation of discrimination performance, DI-HMM can discriminate between different sequences with fewer training data. The error rate of discrimination is less than 0.1 if we train more than two sequences selectively. Therefore, DI-HMM is \changeHKK{powerful} to find \changeHKK{even} the sequence that is not included in the training data. \changeHKK{This feature} {\changeHK{allows} us to \changeHK{easily add new labels into} \changeHK{existing d}atabase\changeHK{s} of the training data.} Furthermore, the evaluation results obtained using the sound data show that DI-HMM \changeHK{gives} higher performances for rhythm pattern recognition than HSMM does by taking \changeHKK{into} account the slight time delay. Therefore, \changeHKK{we can say that} DI-HMM \changeHKK{supports temporal order as well as temporal ambiguity of events to find similar sequential patterns efficiently.}
However, from the evaluation of the calculation time, the proposed method requires additional time to treat the interval. \changeHK{This revealed} the fact that the more additional time might be needed when the number of training data increases.
Future studies will be conducted to compare our proposed method with \changeHK{a \changeHKK{further} new different method} which introduces the interval state node to HSMM,  and to evaluate the training-recognition time and memory consumption. Additionally, we shall improve DI-HMM to reduce such calculation costs to facilitate its application as an online system.

\bibliographystyle{IEEEtran}

\bibliography{IJCNN_arXiv_rev1}

\end{document}